\documentclass[10pt,twocolumn,letterpaper]{article}

\usepackage{iccv}
\usepackage{times}
\usepackage{epsfig}
\usepackage{graphicx}
\usepackage{amsmath}
\usepackage{amssymb}
\usepackage{booktabs}
\usepackage{mathtools, cuted}
\usepackage{caption}

\usepackage{multirow}
\usepackage{colortbl}
\newcommand{\myparagraph}[1]{\vspace{2pt}\noindent{\bf #1}}

\usepackage[pagebackref=true,breaklinks=true,letterpaper=true,colorlinks,bookmarks=false]{hyperref}

\iccvfinalcopy %

\ificcvfinal\fi

\begin{document}

\title{Soundini: Sound-Guided Diffusion for Natural Video Editing}

\author{
Seung Hyun Lee$^1$ \quad Sieun Kim$^1$ \quad Innfarn Yoo$^2$ \quad Feng Yang$^2$  \quad Donghyeon Cho$^1$ \quad Youngseo Kim$^1$ \\
Huiwen Chang$^2$ \quad Jinkyu Kim$^{3*}$ \quad Sangpil Kim$^{1*}$ \\ 
 $^1$Department of Artificial Intelligence and $^3$CSE, Korea University\\
 $^2$Google Research\\
}

\maketitle

\begin{strip}\centering
\vspace{-3.5em}
\includegraphics[width=\textwidth]{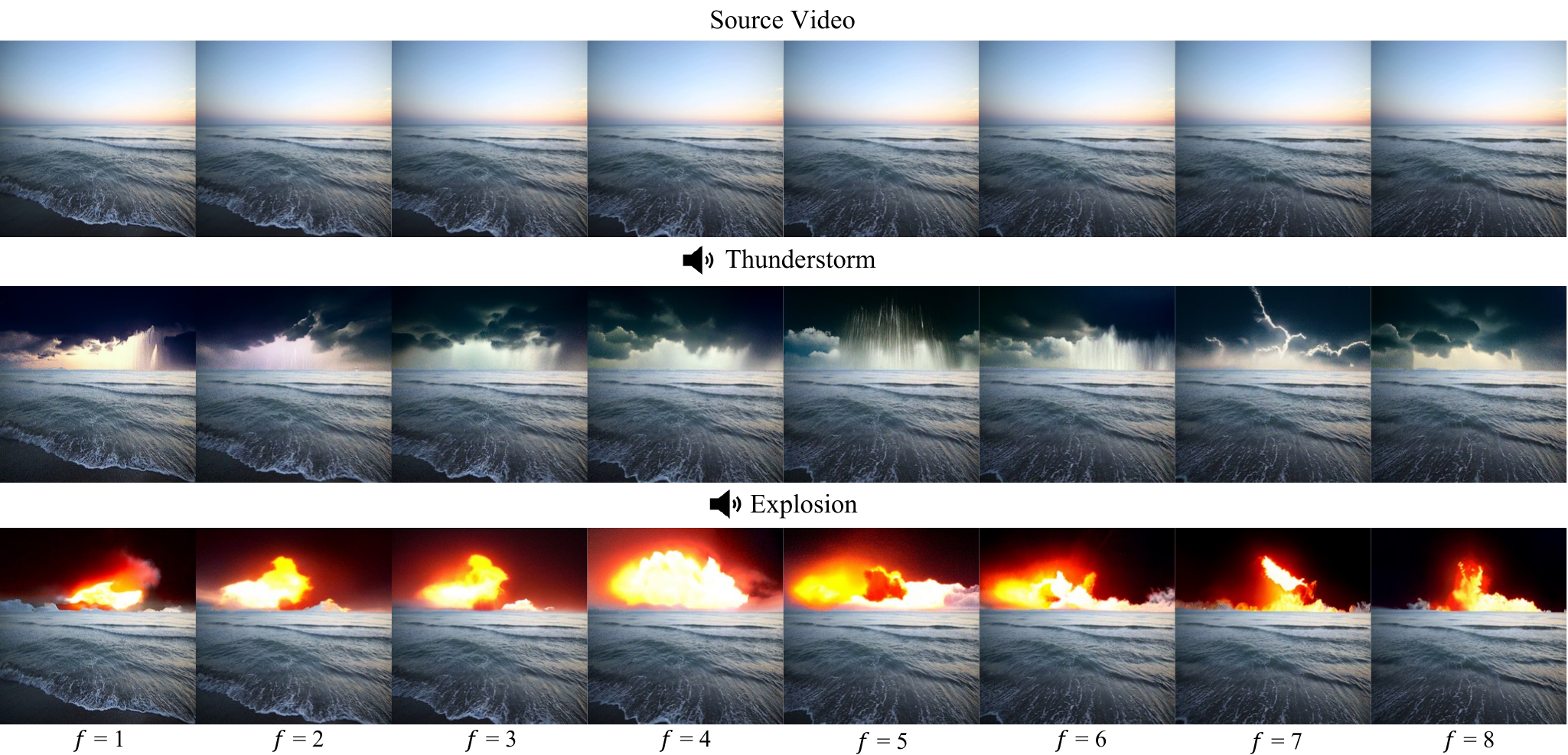}
\captionof{figure}
{
\textbf{Sound-guided video local editing using Soundini} Given a natural video~(top row), our framework edits each frame to be consistent with the meaning of sound in the area of the binary mask. Our model shows realistic fine-detailed visual changes, maintaining temporal consistency. These examples are produced with five frame intervals per column.  
}
\label{fig:fig0}
\vspace{-0.8em}
\end{strip}

\ificcvfinal\thispagestyle{empty}\fi

\begin{abstract}
We propose a method for adding sound-guided visual effects to specific regions of videos with a zero-shot setting. Animating the appearance of the visual effect is challenging because each frame of the edited video should have visual changes while maintaining temporal consistency. Moreover, existing video editing solutions focus on temporal consistency across frames, ignoring the visual style variations over time, e.g., thunderstorm, wave, fire crackling. To overcome this limitation, we utilize temporal sound features for the dynamic style. Specifically, we guide denoising diffusion probabilistic models with an audio latent representation in the audio-visual latent space. To the best of our knowledge, our work is the first to explore sound-guided natural video editing from various sound sources with sound-specialized properties, such as intensity, timbre, and volume. Additionally, we design optical flow-based guidance to generate temporally consistent video frames, capturing the pixel-wise relationship between adjacent frames. Experimental results show that our method outperforms existing video editing techniques, producing more realistic visual effects that reflect the properties of sound. Please visit our page: \url{https://kuai-lab.github.io/soundini-gallery/}.
\end{abstract}

\vspace{-1.4em}
\section{Introduction}
\vspace{-0.5em}

\begin{figure*}[t!]
    \begin{center}
        \includegraphics[width=0.95\textwidth]{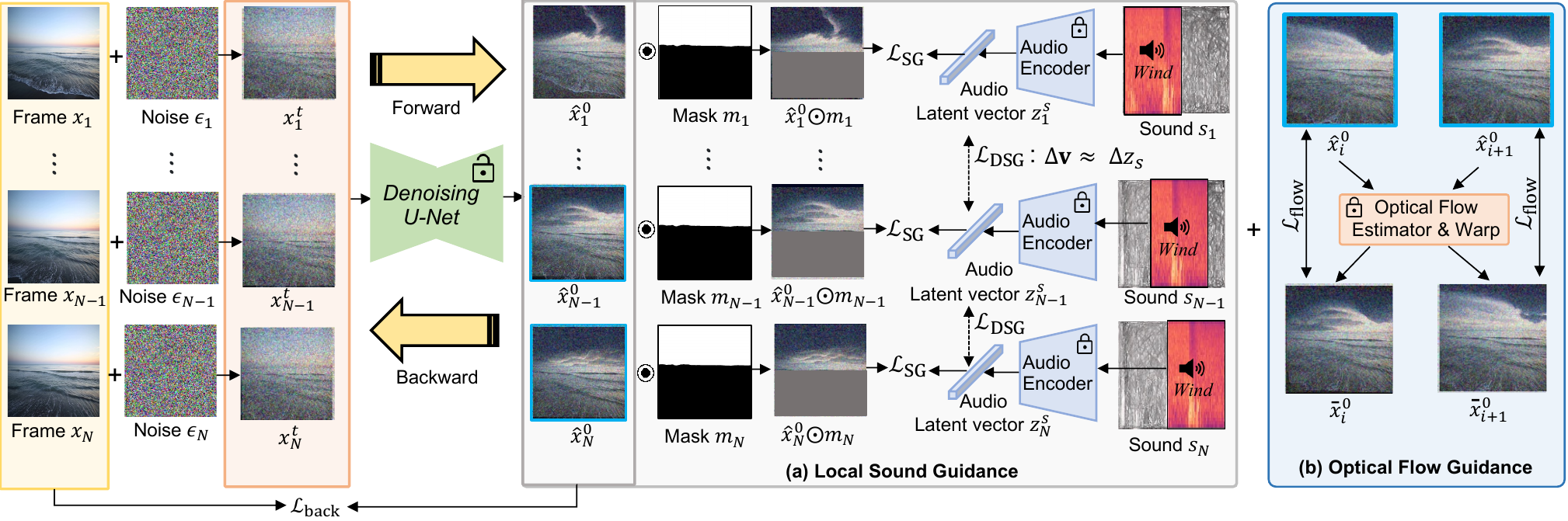}
    \end{center}
    \vspace{-1em}
    \caption{\textbf{Overview of Soundini} Soundini consists of two gradient-based guidance for diffusion: (a)~\textit{Local Sound guidance} and~(b)~\textit{Optical flow guidance}. In (a), we match the appearance of the sampled frame with the sound in the mask region using the loss~$\mathcal{L}_\text{SG}$ and~$\mathcal{L}_\text{DSG}$. First, we minimize the cosine distance loss~$\mathcal{L}_\text{SG}$ between the denoised frame and the sound segments in the audio-visual latent space. Additionally, the loss term $\mathcal{L}_\text{DSG}$ aligns the variation of frame latent representation and that of sound, capturing the transition of sound semantics and reaction. In~(b), our optical flow guidance $\mathcal{L}_\text{flow}$ allows the sampled video to maintain temporal consistency by measuring the mean squared error between warped and sampled frames using optical flow. In particular, gradually increasing the influence on the loss  $\mathcal{L}_\text{flow}$ helps to sample the temporally consistent video stably. Background reconstruction loss $\mathcal{L}_\text{back}$ allows the background to be close to the source video.}
    \label{fig:overview}
    \vspace{-1.0em}
\end{figure*}

Video editing is essential in computer vision fields, with the remarkable development of practical applications in movie-making and social media content creation. Recent user-aided visual editing tools~\cite{brown2022end,chang2023muse,sofyan2020digital} have made it possible to add artistic or realistic visual effects, such as drizzling or explosions, to images and videos by rendering repetitive patterns. However, editing each frame manually requires significant effort to produce a temporally coherent video. 

There are two main approaches for automatic video editing: video decomposition-based and generative models based. The former decomposes the representation of a specific object and background, then edits the appearance of the objects and re-renders to video~\cite{bar2022text2live,kasten2021layered}. Despite their outstanding temporal consistency, these methods can edit only a specific object, not an arbitrary area of the video. The other approach utilizes GAN models pre-trained with large-scale, high-quality images to edit the semantic attributes of each frame over the video~\cite{alaluf2023third,xu2022temporally}. However, both approaches are limited to producing only static styles and struggle to add dynamic visual effects.

We utilize the ability of sound to represent the dynamic scene context through factors specialized in audio, such as timbre, intensity, and volume. Existing works~\cite{lee2022robust,lee2022sound2,lee2022sound} show the usefulness of sound in generating and manipulating visual content by leveraging Generative Adversarial Networks~(GANs)~\cite{brock2018large}. However, when editing real videos, each frame of the video must be projected to the latent space of the GANs~\cite{abdal2019image2stylegan,abdal2020image2stylegan++,alaluf2021restyle,alaluf2022hyperstyle}. This process involves a trade-off between the quality of image reconstruction and the edit-ability~\cite{tov2021designing}.
Therefore, we explore the potential power of sound for natural video editing using Denoising Diffusion Probabilistic Models~(DDPMs)~\cite{dhariwal2021diffusion}. DDPMs have emerged as a powerful architecture for generating high-resolution images from Gaussian noise and can capture fine-detailed knowledge~\cite{liu2023more,rombach2022high}. However, adding visual effects to video with DDPMs remains challenging due to the absence of prior motion knowledge between adjacent frames.  

To overcome these problems, we propose a novel framework that takes advantage of the acoustic characteristics of the sound input. Our framework utilizes a guidance-based diffusion sampling strategy incorporating a pre-trained audio latent representation to produce sound-guided motion and appearance editing. Furthermore, rather than denoising each video frame independently, optical flow-based guidance  allows sampled adjacent frames to keep temporal consistency by matching warped frames to each other using estimated optical flow. 

Experimental results show that our method can produce realistic video editing from various sound sources. For example, a video of the ocean is edited into the ocean with a thunderstorm-like exterior appearance using the sound of thunderstorm~(see Fig.~\ref{fig:fig0}). We also demonstrate that our procedure produces a temporally consistent video without further training in the video dataset. 
Our main contributions are listed as follows:
\begin{itemize}
    \item We propose a novel local sound-guided diffusion for video editing with a sound and user-provided mask. In particular, our method can synthesize a visual effect that reflects properties of sound, such as temporal context, intensity, and volume. 
    \vspace{-0.5em}
    \item We develop an optical flow guidance leveraging an image-based diffusion model. We demonstrate that this procedure is useful for obtaining temporal consistency.
    \vspace{-0.5em}
    \item In terms of style animating, we achieve state-of-the-art performance compared to existing video editing methods, improving perceptual realism. 
\end{itemize}

\vspace{-1.2em}
\section{Related Work}
\vspace{-0.5em}

\myparagraph{Video Editing} For conventional video editing methods, achieving temporal consistency is crucial to ensure naturalness and perceptual realism by preventing temporal jittering and loss of content. Previous GAN-based works~\cite{chen2017coherent,huang2016temporally} have attempted to enforce constraints with optical flow between adjacent frames without further training on the video dataset. Additionally, recent works~\cite{alaluf2023third,jiang2023identity,seo2022styleportraitvideo,wang2023towards,xu2022temporally,xu2023n,yang2022vtoonify} edit each frame by leveraging StyleGAN~\cite{karras2021alias,karras2019style} with semantic disentanglement. However, these works focus on editing face, and the results are limited to the domain of the GAN latent space.

For open-domain video editing, several works~\cite{bar2022text2live} edit the video with a user-provided text prompt using pre-trained image-text joint latent representation. Despite their remarkable performance in video editing, they cannot produce movable visual effects because these works focus only on the contents of consistent objects. On the other hand, our framework can generate more realistic video editing results with movable stylization. Additionally, recent autoregressive approaches~\cite{liang2022nuwainfinity,wu2022nuwa1} propose video manipulation according to given text inputs. However, those works fail to generate dynamic motion because their works highly depend on the order of visual patches. 

Recently, text-driven video diffusion models~\cite{molad2023dreamix,wu2022tune} show successful motions and appearance editing for videos. On the other hand, our method has the ability to control continuous visual changes with audio manipulation.

\begin{figure}[t!]
    \begin{center}
        \includegraphics[width=\linewidth]{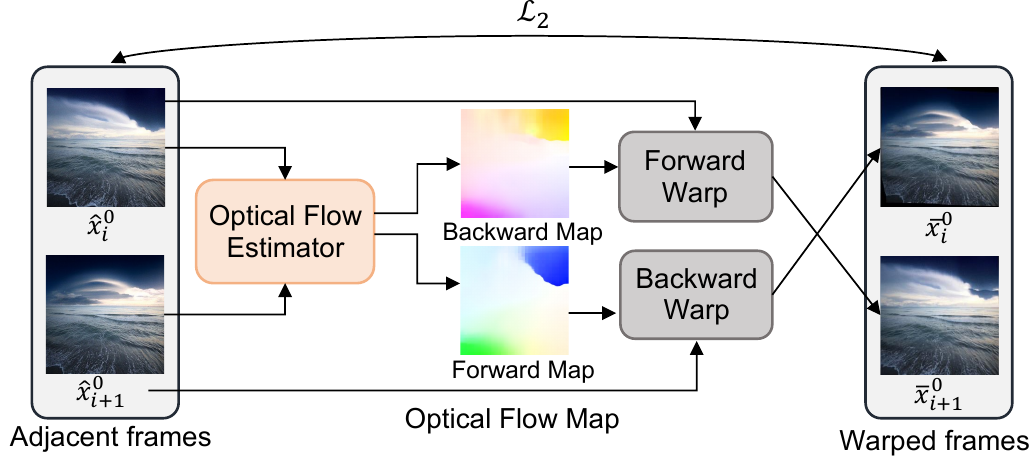}
    \end{center}
    \vspace{-1em}
    \caption{\textbf{Illustration of optical flow guidance for diffusion} The frozen optical flow estimator produces bi-directional optical flows between the sampled frame $i$ and frame $i+1$. Then, we apply forward and backward warp to match each pixel and minimize the mean squared error between the adjacent and the corresponding warped frames.}
    \label{fig:flow}
    \vspace{-1em}
\end{figure}

\myparagraph{Diffusion Models}
Given text descriptions, recent diffusion models~\cite{avrahami2022blended,cheng2023adaptively,Gu_2022_CVPR,ho2020denoising,kawar2022imagic,Kim_2022_CVPR,nichol2021glide,Rombach_2022_CVPR,zhang2022sine} have shown remarkable success in visual editing tasks. From randomly sampled gaussian noise, DDPMs~\cite{dhariwal2021diffusion}, a class of probabilistic generative models, has the ability to denoise the image from noise iteratively~\cite{rombach2022high}, which exceeds the image generation quality of state-of-the-art GAN~\cite{dhariwal2021diffusion}. In particular, blended diffusion models~\cite{avrahami2022blended} proposes an architecture for region-based image editing with text guidance. Furthermore, gradient-based diffusion guidance techniques~\cite{ho2021classifierfree,nichol2021glide} effectively control the appearance and style of the image without further training during image sampling. However, their sampling process is inappropriate for generating temporally consistent videos because they do not consider any relationship between frames. To solve the issue mentioned above, we propose an optical flow guidance diffusion to achieve temporal consistency.

\myparagraph{Sound-Driven Visual Synthesis} Recent works~\cite{lee2022sound2,li2022learning} have shown the effectiveness of sound in changing the fine-detailed style and appearance of objects in an image or video. Li~\textit{et al.}~\cite{li2022learning,zelaszczyk2022audio} proposes to learn visual styles solely based on audio information, leading to anticipated changes in the image style according to volume or mixed audio. In addition, the recent success of learning audio latent representation~\cite{guzhov2022audioclip,jonason2022timbreclip,sheffer2022hear,wu2022wav2clip} shows that the joint audio-visual latent space is helpful for sound-guided visual synthesis. 

Among them, Lee~\textit{et al.}~\cite{lee2022sound} also takes advantage of sound properties, such as intensity, as well as semantic cues for sound-guided semantic image manipulation. Furthermore, the work shows high-resolution image synthesis with audio using CLIP~\cite{radford2021learning} and StyleGAN latent space~\cite{karras2019style}. However, these works only focus on sound semantic cues and are not able to generate audio-relevant motion. In contrast, we match the motion of visual effects with the variation of the audio signal during the diffusion frame sampling process. MM-Diffusion~\cite{ruan2022mm} generates audio-video simultaneously, but it requires audio-visual pairs during training.

\begin{figure*}[t!]
    \begin{center}
        \includegraphics[width=0.95\textwidth]{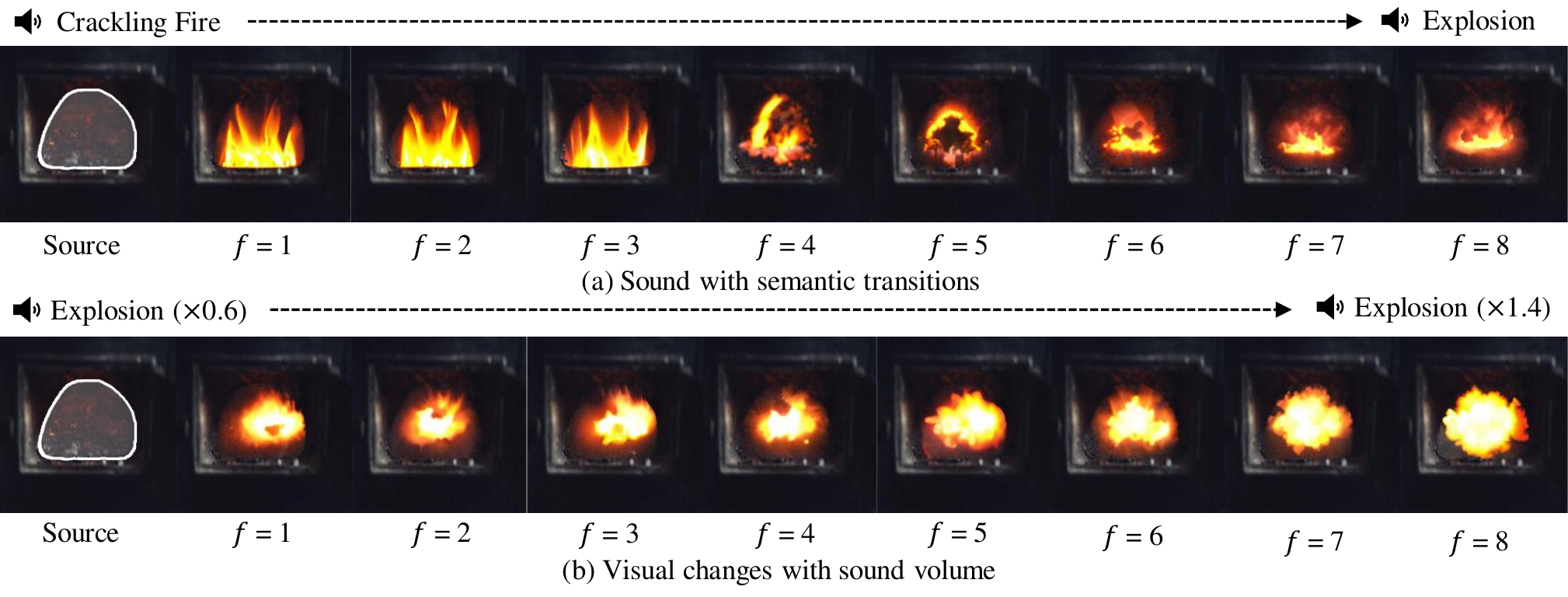}
    \end{center}
    \vspace{-1.5em}
    \caption{\textbf{Effect of sound} (a)~Sounds with semantic visual transitions. Given the video and user-provided mask, we use the sound input with two distinctive meanings,~\textit{Crackling fire} and~\textit{Explosion}. We observe that visual changes are smooth corresponding to sound semantics. (b)~Visual changes with sound volume. As we increase the magnitude of sound~\textit{Explosion}, the style of the image also becomes drastic. There are five intervals per column. }
    \label{fig:transition}
    \vspace{-1.0em}
\end{figure*}

\vspace{-1.2em}
\section{Method}
\vspace{-0.5em}
We introduce a novel framework for sound-guided video local editing~(see Fig.~\ref{fig:overview}). First, we take a one-dimensional audio waveform as input and transform the audio waveform into mel spectrogram. Given the preprocessed mel spectrogram sound input $s \in \mathbb{R}^{M \times L}$~(2D), binary mask $m \in \mathbb{R}^{N \times 1 \times W\times H}$, and source video $x\in \mathbb{R}^{N \times C \times W\times H}$, we predict the edited video $\hat{x}^{N \times C \times W\times H}$, where $N$ is the number of frames, $W, H, C$ denote the width, height, and channel, respectively, $M$ and $L$ are the number of mel spectrogram filters and the time duration. 

Our goal is to make visual changes consistent with the sound in the region of interest. To achieve this goal, our model presents two main guidance for diffusion sampling:~(i) Local Sound Guidance and (ii) Optical Flow Guidance. Section~\ref{method:sound} explains how local sound guidance diffusion matches each frame with semantic cues and temporal variation of sound chunks. In Section~\ref{method:optical}, we propose bidirectional optical flow guidance for achieving temporal consistency across sampled frames.

\vspace{-0.5em}
\subsection{Local Sound Guidance}
\vspace{-0.5em}
\label{method:sound}

\myparagraph{DDPMs for Video Editing} 
Our proposed method for sound-guided natural video editing utilizes DDPMs~\cite{dhariwal2021diffusion}. To generate a forward Markov chain $x_1,...,x_T$, DDPMs obtain $x_t$ by adding noise to the clean image $x_0$ at the time step $t$. Note that the last image $x_T$ approaches a Gaussian distribution, particularly in the case of large $T$. 

To extend this approach to video, the forward Markov chain $x_i^1,...,x_i^T$ is operated independently for each frame where $i$ denotes the frame index for $i \in \{1,2,...,N\}$ and $N$ denotes the frame number of the source video. Given a data distribution $x_i^0 \sim q(x_i^0)$ of the frame $i$, we define the Markov transition $q(x_i^t| x_i^{t-1})$  from the normal distribution using the variance schedule $\beta_t \in (0,1)$ as follows:
\begin{equation}
    \begin{aligned}
        q(x_i^t| x_i^{t-1}) = \mathcal{N}(x_i^t;\sqrt{1-\beta_t}x_i^{t-1}, \beta_t \mathbf{I}), \quad t=1,...,T.
    \end{aligned}
    \label{eq:ddpm1}
\end{equation}

Without an intermediate step, the forward noising process produces $x_i^t$  from $x_i^0$ by adding noise as follows:
\begin{equation}
    \begin{aligned}
        q(x_i^t| x_i^{0}) = \mathcal{N}(x_i^t;\sqrt{1-\beta_t}x_i^{0}, \beta_t \mathbf{I}) \\
        x_i^t=\sqrt{\bar{\alpha_t}} + \sqrt{1 - \bar{\alpha_t}} \epsilon_i,
    \end{aligned}
    \label{eq:ddpm1_1}
\end{equation}
where $\epsilon_i \sim \mathcal{N}(0, \mathbf{I})$ and $\alpha_t = 1-\beta_t$, and $\bar{\alpha}_t=\prod_{s=1}^t \alpha_s$. In order to denoise and generate the Markov chain, we utilize the reverse process with a prior distribution of $p(x_i^T)=\mathcal{N}(x_i^T;0,\mathbf{I})$. Estimating the parameters, $\theta$ ensures that the generated reverse process matches closely the fixed forward process as follows:
\begin{equation}
    \begin{aligned}
        p_\theta(x_i^{t-1} | x_i^t) = \mathcal{N}(x_i^{t-1};\mu_\theta(x_i^t, t), \Sigma_{\theta} (x_i^t, t)), 
    \end{aligned}
    \label{eq:ddpm2}
\end{equation}
where $t=T,...,1$ denotes the time step of the reverse process. Followed by Ho~\textit{et al.}~\cite{ho2020denoising}, we allow the denoising autoencoder $\epsilon_\theta(x_i^t, t)$ to denoise $x_i^T$ which is a form of noise added $x_i^0$ instead of directly estimating $\mu_\theta(x_i^t, t)$ by Bayes' theorem as follows:
\begin{equation}
    \begin{aligned}
        \mu_\theta(x_i^t, t) = {1 \over \sqrt{\alpha_t}} (x_i^t - {\beta_t \over \sqrt{1-\bar{\alpha}_t}} \epsilon_\theta(x_i^t, t)).
    \end{aligned}
    \label{eq:ddpm3}
\end{equation}

Therefore, the denoising autoencoder $\epsilon_\theta(x_i^t, t)$ predicts the denoised frame $x_i^0$ from each noisy latent diffusion $x_i^t$. Given $x_i^t$ and time step $t$, the predicted $i$-th frame $\hat{x}_i^0$ is directly obtained from $\epsilon_\theta(x_i^t, t)$ at each time step as follows:
\begin{equation}
    \begin{aligned}
        \hat{x}_i^0 = {x_i^t - \sqrt{1-{\bar{\alpha}}_t}\epsilon_\theta(x_i^t,t) \over \sqrt{\bar{\alpha}_t}}.
    \end{aligned}
    \label{eq:sound}
\end{equation}

\myparagraph{Local Sound Guidance} Blended-diffusion~\cite{avrahami2022blended} leverages image-text multi-modal embedding space pre-trained with a large-scale image and text pairs to guide diffusion towards a target prompt. However, we extend this approach to video by averaging the loss of cosine distance of each frame with a given audio fragment $s_i$. To achieve this, we sample frames by sliding the mel spectrogram $s_i$ along the time axis and extract $d$-dimensional sound latent representation from given sound chunks $s_i$. Note that the audio encoder $f_s(\cdot)$ makes our sound latent representation semantically consistent with the latent image representation. We measure the cosine distance between the normalized latent vector as follows: 
\begin{equation}
    \begin{aligned}
        \mathcal{L}_\text{SG} = {1 \over N} \sum_{i=1}^N d_\text{cos}(f_v(\hat{x}_{i}^0 \odot m_i), f_s(s_i)),
    \end{aligned}
    \label{eq:clip}
\end{equation}
where $d_\text{cos}(\cdot, \cdot)$ denotes the cosine distance and $f_v(\cdot)$ denotes the CLIP image encoder. Furthermore, $\odot$ denotes pixel-wise multiplication, and $m_i$ denotes the binary mask according to each frame.

We not only focus on matching the appearance and style of objects with the meaning of sound but also on visual changes corresponding to sound transitions. To generate sound-related motion for each frame, we minimize the cosine distance between the direction of audio embedding $z_s$ and the direction of frame embedding between adjacent frames as follows:
\begin{equation}
    \begin{aligned}
        \mathcal{L}_\text{DSG} = {1 \over {N-1}} \sum_{i=1}^{N-1} d_\text{cos}(\Delta_{i}^{i+1} \textbf{v}, \Delta{z_s}),
    \end{aligned}
    \label{eq:sound}
\end{equation}
where $d_\text{cos}$ is the cosine distance between the latent representation of the frame and the audio. The direction of the visual latent representation $\Delta_{i}^{i+1} \textbf{v}$ is calculated as $f_v(\hat{x}_{i+1}^0 \odot m_{i+1}) - f_v(\hat{x}_{i}^0 \odot m_i)$. In this way, we effectively produce movable visual effects corresponding to sound signals by measuring the variation of image embedding.

\begin{figure}[t!]
    \begin{center}
    \includegraphics[width=\linewidth]{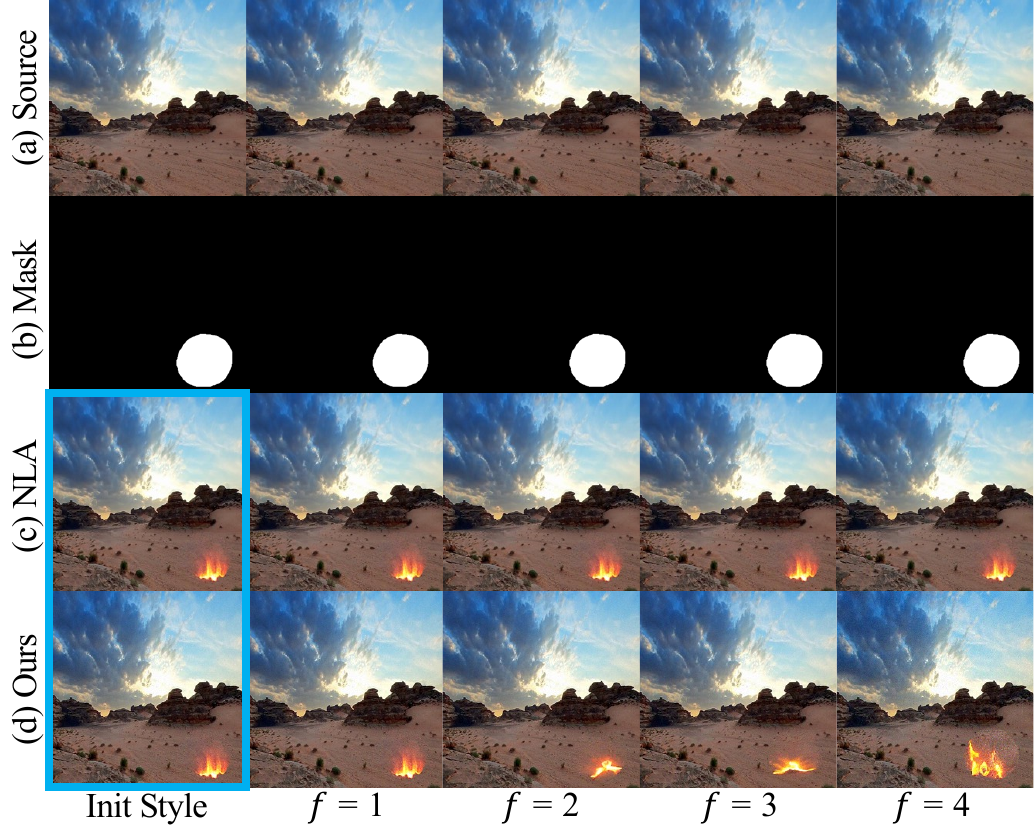}
    \end{center}
    \vspace{-1em}
    \caption{\textbf{Comparison with local video editing} Given the (a) source video, we visualize the edited video with~\textit{crackling fire} sound and the (b) binary mask, which is annotated manually. We compare~(d) our method with~(c) Neural Layered Atlases~(NLA)~\cite{kasten2021layered}~(five frame intervals for each column). We warp the same localized style for the first frame. }
    \label{fig:result1}
    \vspace{-1.1em}
\end{figure}

\vspace{-0.5em}
\subsection{Optical Flow Guidance}
\vspace{-0.5em}
\label{method:optical}
\myparagraph{Optical Flow Guidance} 
To achieve temporal consistency, we additionally provide optical flow-based guidance during the diffusion process~(see Fig.~\ref{fig:flow}). First, we obtain a bidirectional optical flow from the sampled frame $i$ and the sampled frame $i+1$ with RAFT~\cite{teed2020raft}, the pre-trained optical flow estimation network. Next, we warp each frame $\hat{x}_{i}^0$, $\hat{x}_{i+1}^0$ into the new warped frames $\bar{x}_{i}^0$, $\bar{x}_{i+1}^0$ with estimated optical flow. Given the source video of $N$ frames during the diffusion process, we compute the mean squared loss between the frames as follows:
\begin{equation}
    \begin{aligned}
        \mathcal{L}_\text{flow} = {1 \over N-1} \sum_{i=1}^{N-1} (\mathcal{L}_2(\hat{x}_{i}^0, \bar{x}_{i}^0) + 
        \mathcal{L}_2(\hat{x}_{i+1}^0, \bar{x}_{i+1}^0)).
    \end{aligned}
    \label{eq:flow}
\end{equation}
This term ensures the pixel value across sampled frames from the noisy version of the source frame. We apply optical flow guidance to the global rather than the local area because temporal coherent visual effects can be created considering the global context.

\begin{figure}[t!]
    \begin{center}
        \includegraphics[width=\linewidth]{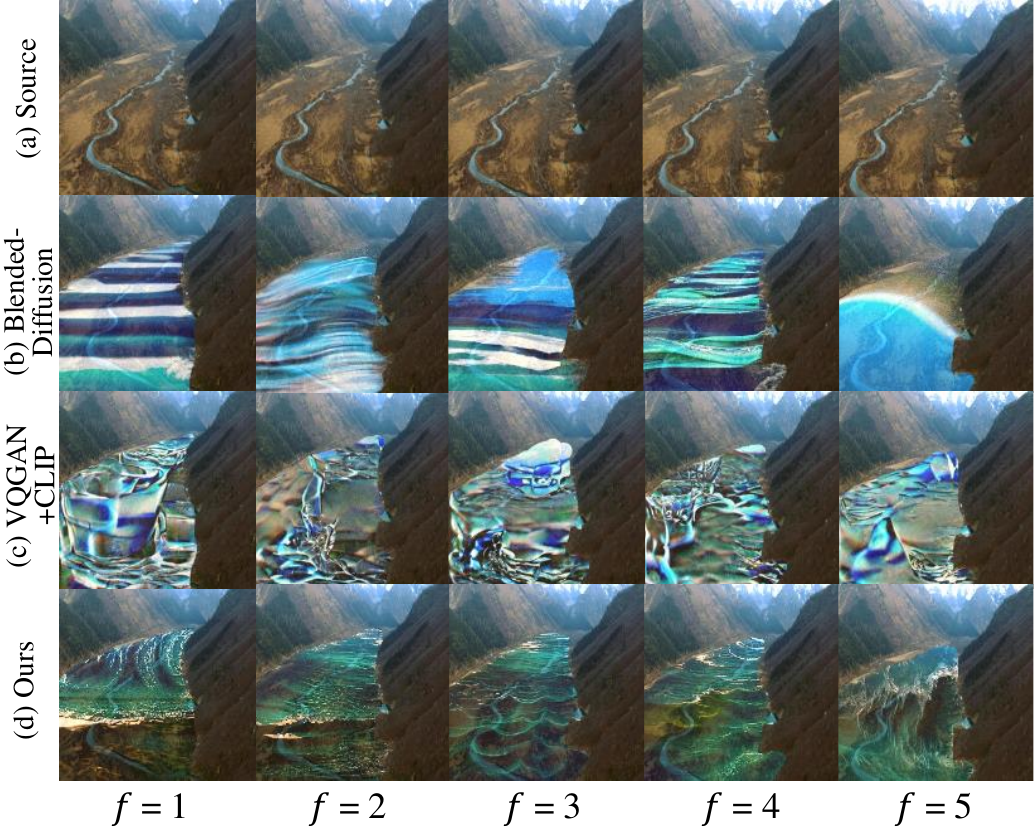}
    \end{center}
    \vspace{-1em}
    \caption{\textbf{Comparison with the extension of image editing works} Given~\textit{Wave} sound and manually annotated binary mask, we edit the~(a) input video with other baselines, (b) Blended Diffusion~\cite{avrahami2022blended}, (c) VQGAN+CLIP~\cite{crowson2022vqgan}, and (d) ours respectively~(five frame intervals per column). }
    \label{fig:result2}
    \vspace{-1.1em}
\end{figure}

\myparagraph{Background Reconstruction Loss} Background reconstruction is also an important issue in terms of producing seamless results. To maintain the background of source frames, we apply the background preserve guidance as follows: 
\begin{equation}
    \begin{aligned}
        \mathcal{L}_\text{back} = {1 \over N} \sum_{i=1}^N \mathcal{L}_1(\hat{x}_{i}^0 \odot (1 - m_i), {x}_{i}^0 \odot (1 - m_i)) + \\ \mathcal{L}_\text{LPIPS}(\hat{x}_{i}^0 \odot (1 - m_i), {x}_{i}^0 \odot (1 - m_i)),
    \end{aligned}
    \label{eq:back}
\end{equation}
where $\mathcal{L}_1$ and $\mathcal{L}_\text{LPIPS}$ denote the loss of L1 in pixels and the loss of perceptual~\cite{johnson2016perceptual}, which captures fine-detailed features of the background of the source frame. $\odot$ denotes pixel-wise multiplication. After sampling, we substitute the mask area with the matching area from the input frame, thereby maintaining the background.

\myparagraph{Total Guidance Loss} To obtain the temporal consistency of the edited video, 
\begin{equation}
    \begin{aligned}
        \mathcal{L}_\text{total} = \lambda_\text{SG}\mathcal{L}_\text{SG} + \lambda_\text{DSG}\mathcal{L}_\text{DSG} + {T - t \over T}\lambda_\text{flow}\mathcal{L}_\text{flow} \\ +  \lambda_\text{back}\mathcal{L}_\text{back},
    \end{aligned}
    \label{eq:total}
\end{equation}
where $\{\lambda_\text{SG}, \lambda_\text{DSG}, \lambda_\text{flow}, \lambda_\text{back}\}$ are a set of hyperparameters to determine the power of each term. Especially, we adjust $\lambda_\text{flow}$ to control the trade-off relationship between temporal consistency and dynamic motions. We gradually increase $\lambda_\text{flow}$ depending on time step $t$ because the sampled frames are almost unrecognizable when $t$ is near to $T$. Then, we sample noisy latent diffusion $x^{t-1}_i$ at time step $t-1$ as follows: 
\begin{equation}
    \begin{aligned}
        x^{t-1}_i \sim \mathcal{N}(\mu + \Sigma\nabla_{\hat{x}^0_i}\mathcal{L}_\text{total}, \Sigma),
    \end{aligned}
    \label{eq:total}
\end{equation}
where $\mu$ and $\Sigma$ indicate $\mu_\theta(x_i^t, t)$ and $\Sigma_\theta(x^t_i, t)$ respectively.

\begin{figure}[t!]
    \begin{center}
    \includegraphics[width=\linewidth]{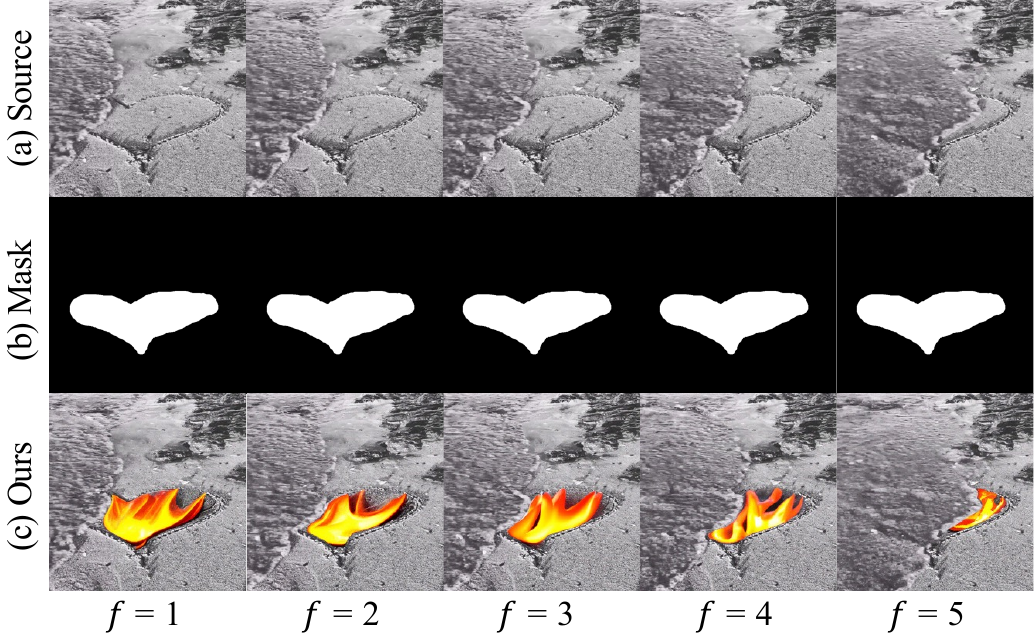}
    \end{center}
    \vspace{-1em}
    \caption{\textbf{Context adaptive video editing} The first row is the source video, the second row is the mask, and the third row is the edited results of ours. Given the fire crackling sound, we add reasonable visual effects within the local area, considering the global context of the original video. }
    \label{fig:mask}
    \vspace{-1em}
\end{figure}

\vspace{-0.5em}
\section{Result}
\vspace{-0.5em}

In this section, we discuss the details of our implementation and experiments. 
\vspace{-0.5em}
\subsection{Implementation Details}
\vspace{-0.5em}

\myparagraph{Evaluation Metrics} We quantitatively compare our method to baselines on the DAVIS~\cite{pont20172017} dataset, which provides the high-resolution video and the ground truth binary mask manually annotated per frame. For fairness, we resize the videos at 256$\times$256 resolution and sample 10 frames. To evaluate the temporal consistency of the video editing results, we adopt the conventional temporal metric, T-diff~\cite{chen2017coherent}, which calculates the pixel-wise difference of warped frames. For evaluating video quality, we adopt three widely acceptable quantitative metrics: Fr\'echet Video Distance~(FVD)~\cite{unterthiner2019fvd}, Fr\'echet Image Distance~(FID)~\cite{heusel2017gans}, and Inception Score~(IS)~\cite{NIPS2016_8a3363ab}. In particular, we measure the average of activations across frames because FID and IS are image-based metrics. Then, we leverage the Inception3D~\cite{carreira2017quo} network pre-trained with the Kinetics-400 dataset~\cite{kay2017kinetics} for FVD. To measure semantic consistency, we use the CLIP score~\cite{hessel-etal-2021-clipscore}, where a higher score indicates a stronger semantic alignment between the edited video and the given semantics.

\myparagraph{Hyperparameter Setting} Regarding the optical flow estimation network, we use the pre-trained RAFT~\cite{teed2020raft} with 900 maximum long edges. Each guidance weight $\lambda_\text{SG}, \lambda_\text{DSG}, \lambda_\text{flow}, \lambda_\text{back}$ are set to 1000, 1000, 500, 1000 respectively. When the $\lambda_\text{SG}, \lambda_\text{DSG}$ are set too low, our model fails to match the visual changes with sound. Additionally, with too low $\lambda_\text{flow}, \lambda_\text{back}$, our method cannot produce the seamless results. Total time step $T$ and the sampling computation cost have a trade-off relationship. Enlarging the total time step can lead to the more realistic result. However, the total time step $T$ is set to 100 because we empirically confirm that the improvement in video editing performance is saturated when the total time step is larger than 100~(see supplementary document).

\begin{table*}[t!]   
    \caption{Quantitative comparison between ours and alternatives on the DAVIS~\cite{pont20172017} dataset. We report T-diff~\cite{chen2017coherent} for temporal consistency and FVD~\cite{unterthiner2019fvd}, FID~\cite{heusel2017gans}, IS~\cite{NIPS2016_8a3363ab} for video quality, and CLIP score for semantic consistency. The best score and second-best score are shown in \textbf{bold} and \underline{underlined}. 
    }
    \label{table:all}
    \centering
    \vspace{-0.6em}
    \small
    \begin{tabular}{@{}lccccc@{}}
        \toprule
        \multirow{2}{*}{Model} & \multicolumn{1}{c}{Temporal Consistency} & \multicolumn{3}{c}{Video Quality Metric} & \multicolumn{1}{c}{Semantic Consistency} \\ 
         \cmidrule(lr){2-2}\cmidrule(lr){3-5} \cmidrule(lr){6-6}
        & T-diff~($\downarrow$) & FVD~($\downarrow$) & FID~($\downarrow$)  & IS~($\uparrow$) & CLIP score~($\uparrow$)\\ 
        \midrule
        Neural Layered Atlases~(NLA)~\cite{kasten2021layered} & \textbf{2.2057} & 3649.4945
 & 379.3251 & 5.8814 & 0.5592 \\ 
        \midrule
        Blended Diffusion~\cite{avrahami2022blended} & 35.5234 & 3127.8477
 & 391.1936 & 5.050 & 0.6035 \\ 
        VQGAN + CLIP~\cite{crowson2022vqgan} &  43.8372 &  4939.3658
 & 351.4702 & 4.3484 & 0.5308 \\ 
        \midrule
         Ours w/o $\mathcal{L}_\text{flow}$ & 7.6357 & \underline{3065.4342} & \underline{349.7132} & \underline{6.3487} & \underline{0.6080} \\ 
        Ours &  \underline{5.9172} & \textbf{2718.2134} & \textbf{332.2388} & \textbf{7.1999} & \textbf{0.6138} \\ 
        \bottomrule
    \end{tabular}
    \vspace{-0.8em}
\end{table*}

\begin{figure}[t!]
    \begin{center}
        \includegraphics[width=\linewidth]{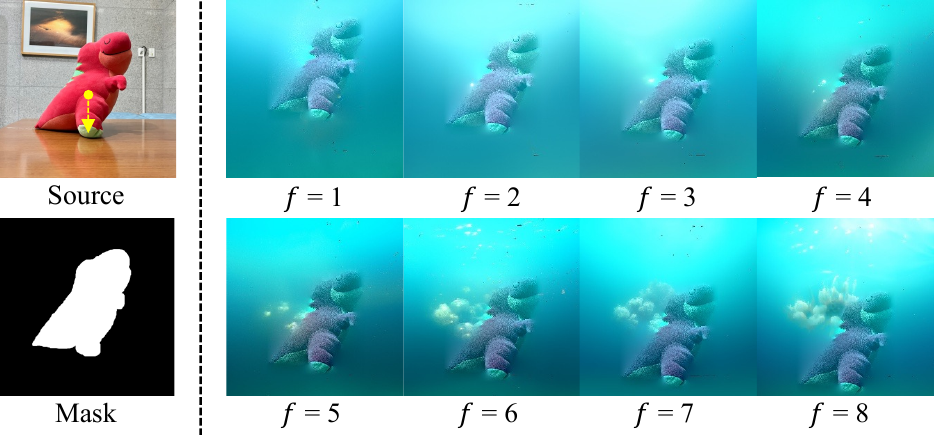}
    \end{center}
    \vspace{-1em}
    \caption{\textbf{Sound-guided video global editing} Our model can edit global areas in the video. Given source image, we change the appearance of the video according to to~\textit{underwater bubbling} sound. We apply widely used image cloning technique~\cite{farbman2009coordinates} for natural blending.}
    \label{fig:joint}
    \vspace{-1.2em}
\end{figure}

\vspace{-0.5em}
\subsection{Qualitative Analysis}
\vspace{-0.5em}
We qualitatively demonstrate the effectiveness of sound in video editing, specifically for semantic transition and volume control. Then, we compare our method to several prominent video editing methods. To the best of our knowledge, we propose the first work that performs local natural video editing with given natural sound inputs. Since there are no perfectly identical settings, we compare two types of baseline: local video editing and image editing extension. We emphasize that our comparison experiments are conducted solely based on sound. Additionally, we provide the video in supplemental materials. 

\vspace{-1.2em}
\subsubsection{Effectiveness of Sound in Video Editing}
\vspace{-0.8em}
\myparagraph{Semantic Transition using Sound} Since sound contains any semantic changes in acoustic features over time, we can produce an edited video with the corresponding sound variation. As Fig.~\ref{fig:transition}~(a) shows, our schemes can guide diffusion models in this challenging setting, sound transition. Changing from~\textit{Cracking fire} sound to~\textit{Explosion} sound, we observe that our framework can capture the semantic transition solely based on the acoustic features of the input sound, and the semantics of the video is still consistent with the sound. This is because Eq.~\ref{eq:sound} is capable of providing guidance regarding the variation of visual latent representation without any latent interpolation.

\myparagraph{Visual Changes with Sound Volume} Sound has one of the unique properties, volume, which is beneficial in deciding the magnitude of visual changes. By increasing the waveform scale, Fig.~\ref{fig:transition}~(b) qualitatively shows that our framework is capable of producing visual effects consistent with the sound volume. Given the~\textit{Explosion} sound, the style gradually becomes noticeable as the volume increases.

\begin{figure}[t!]
    \begin{center}
    \includegraphics[width=\linewidth]{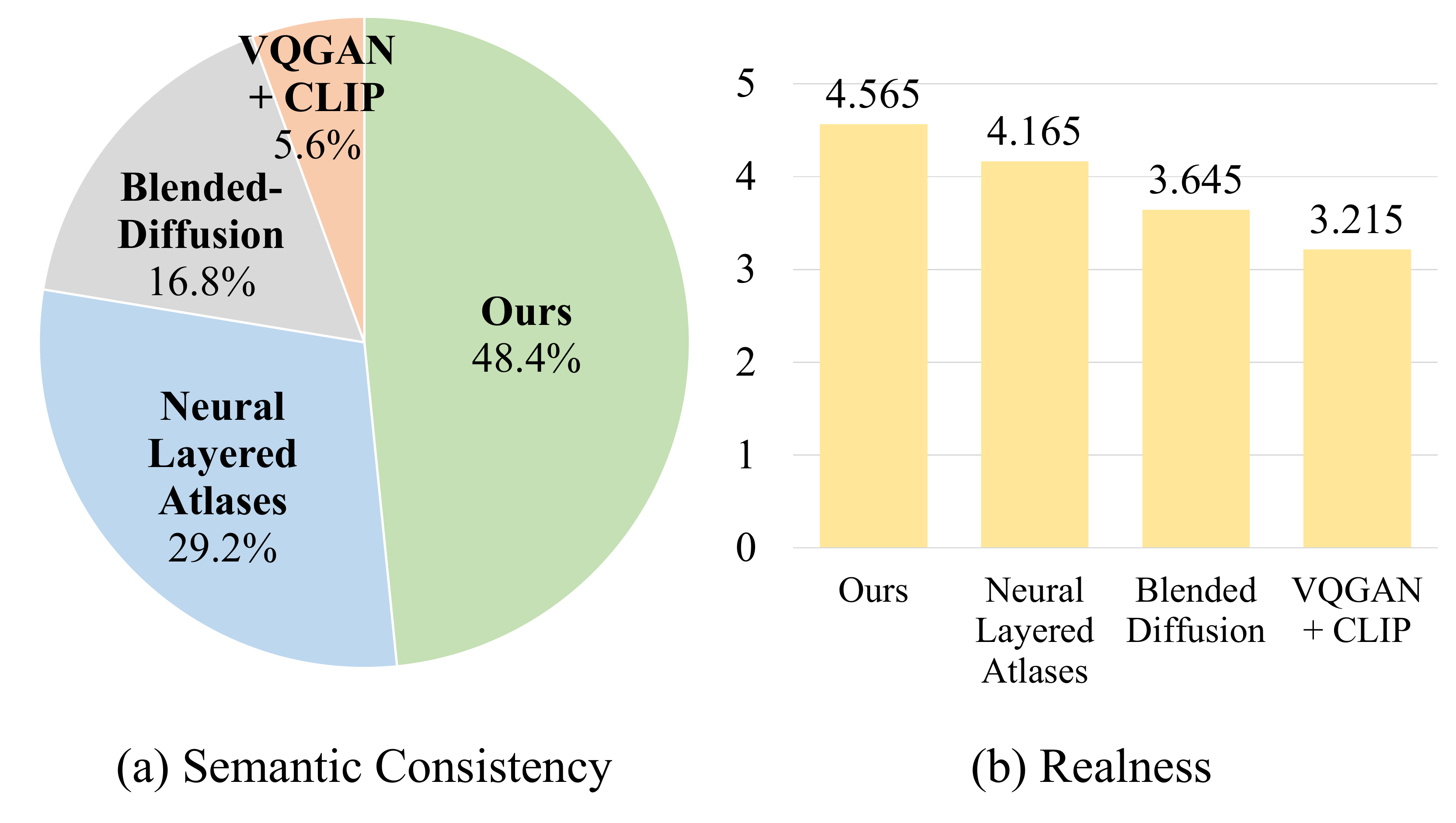}
    \end{center}
    \vspace{-1.8em}
    \caption{\textbf{User study between ours and baselines} }
    \label{fig:user}
    \vspace{-1.2em}
\end{figure}

\vspace{-1.2em}
\subsubsection{Comparison with Baselines}
\vspace{-0.8em}

\myparagraph{Localized Video Editing} We compare our method with existing local video editing, Neural Layered Atlases~(NLA)~\cite{kasten2021layered}. As shown in Fig.~\ref{fig:result1}, our framework produces more realistic and natural results compared to NLA. The added visual effect in NLA is static and cannot move, resulting in unnaturally edited videos. Furthermore, the quality of NLA-supported video editing is highly dependent on the quality of video decomposition. In contrast, our approach allows for more flexibility and control over the edited video, as we are able to stylize desired parts of the video with audio input while maintaining temporal consistency. For fairness, we fix the resolution of the frames $256 \times 256$ and set the number of frames at ten. The same visual style guided by sound is set in the first frame of NLA.

\myparagraph{Leveraging Image Generative Models for Video Editing} Our framework can be applied to a similar setting in that we leverage the pre-trained image generative models. Because those works aim to perform text-guided image editing, we use sound latent representation rather than text latent representation for fairness. As Fig.~\ref{fig:result2} shows, our method is more temporally consistent than blended diffusion and VQGAN$+$CLIP~\cite{crowson2022vqgan}. The visual changes produced by blended diffusion fail to maintain geometric properties between frames. Since VQGAN$+$CLIP regularizes the latent vector and uses a discrete latent space for image generation, each frame shares similar fine details of style. However, the motion between frames is not related to the meaning of the text or audio. On the other hand, our framework delivers explicit pixel-level motion, while visual style changes according to sound transition or volume. 

\vspace{-1.5em}
\subsubsection{Context Adaptive Video Editing}
\vspace{-0.5em}
We achieve robust editing performance within the local area by understanding the global context of the video. We let the global context affect the style in the local area. Given~\textit{fire crackling} sound, our results show that the visual effect of the fire is swept away by the waves~(see Fig.~\ref{fig:mask}). We consider the entire context of the video, not just the edited area, so we can get a more natural video editing result.

\vspace{-1.5em}
\subsubsection{Sound Guided Video Global Editing} 
\vspace{-0.5em}
We support sound-guided video global editing, where the entire frame is regarded as the foreground. After sampling, we blend the sampled frames and the source video using the seamless image cloning technique~\cite{farbman2009coordinates} while preserving the texture and lighting of the target video. Our framework can produce the video of the object slowly sinking in water according to~\textit{underwater bubbling} sound~(see Fig.~\ref{fig:joint}).

\vspace{-0.5em}
\subsection{Quantitative Evaluation}
\vspace{-0.5em}

We quantitatively compare our method with three baselines: Neural Layered Atlases~(NLA)~\cite{kasten2021layered}, Blended Diffusion~\cite{avrahami2022blended}, and VQGAN$+$CLIP~\cite{crowson2022vqgan} on the DAVIS~\cite{pont20172017} dataset in a zero-shot setting. We use ten types of audio samples, including~\textit{fire crackling, explosion, thunderstorm, raining, underwater bubbling}, etc~(see Table~\ref{table:all}). 

\myparagraph{Temporal Consistency} We observe that optical flow guidance leads to higher temporal consistency than other baselines, except NLA. This is because the NLA applies geometric transformations of the visual style without any changes in fine-grained details.

\begin{figure}[t!]
    \begin{center}
    \includegraphics[width=\linewidth]{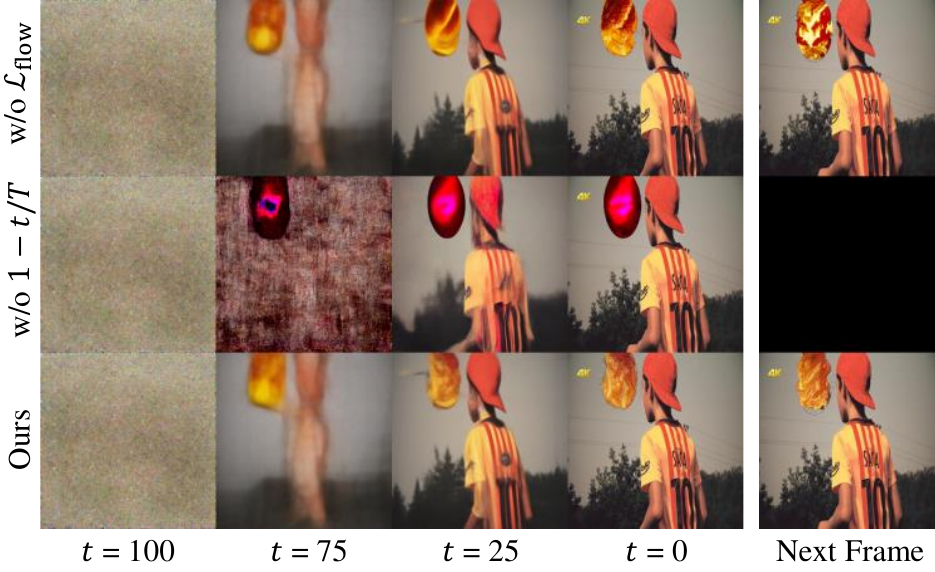}
    \end{center}
    \vspace{-1.2em}
    \caption{\textbf{Ablation study of optical flow guidance} The first shows the sampling results without $\mathcal{L}_\text{flow}$. The second row shows the sampling results without adjusting $\mathcal{L}_\text{flow}$. The third shows the sampling results with increasing $\mathcal{L}_\text{flow}$ after applying optical flow-based guidance.}
    \label{fig:ablation}
    \vspace{-1.5em}
\end{figure}

\myparagraph{Video Quality} Although Blended Diffusion and VQGAN$+$CLIP generate high-quality frames individually, our method obtains higher quality video than baselines. Furthermore, we found that our method with $\mathcal{L}_\text{flow}$ produces better results compared to the method without it, indicating that $\mathcal{L}_\text{flow}$ is effective in ensuring the video quality. This is because image-generative models sample each frame independently, but ours regulate the variety of styles with optical flow-based guidance. 

\myparagraph{CLIP score} We measure the cosine similarity score with the CLIP version \texttt{ViT-B/32}~\cite{radford2021learning} between the edited video and text labels. CLIP image encoder extracts 512-dimensional video embeddings corresponding to each frame and averages them. In addition, 512-dimensional text embeddings are also extracted from the CLIP text encoder. We illustrate that our method achieves the best semantic consistency compared to any other baseline, which means that edited videos are more visually correlated with sound. 

\vspace{-0.5em}
\subsection{User Study}
\vspace{-0.5em}
We conduct a human evaluation study by assembling 100 participants from Amazon Mechanical Turk~(AMT). Participants are asked to choose between the edited results of Neural Layered Atlases~(NLA)~\cite{kasten2021layered}, Blended Diffusion~\cite{avrahami2022blended}, VQGAN$+$CLIP~\cite{crowson2022vqgan}, and ours from 20 questions. We investigate two factors: Semantic Consistency~(i)~\textit{Which video editing results are better consistent with the target attribute?} and Realness~(ii)~\textit{Please evaluate how realistic the video is}. We evaluate perceptual realism by measuring the Likert scale from 1~(low realistic) to 5~(high realistic). Fig.~\ref{fig:user} shows that Soundini outperforms the other methods regarding semantic consistency and realness.

\vspace{-0.5em}
\subsection{Ablation Study}
\vspace{-0.5em}
Fig.~\ref{fig:ablation} shows the effect of optical flow guidance in terms of temporal consistency. We compare our method with the removed version of the specific loss $\mathcal{L}_\text{flow}$ in Eq.~\ref{eq:flow} and the removed version of scheduling $1 - {t\over T}$ in Eq.~\ref{eq:total}. In addition, we visualize the denoised frame of each time step between adjacent frames. We observe that the use of optical flow-based guidance results in higher temporal consistency compared to the version without it. In contrast, independent sampling, without consideration of the relationship, fails to maintain temporal consistency. This demonstrates the effectiveness of the proposed optical flow-based guidance in improving the temporal consistency of our method.

\begin{figure}[t!]
    \begin{center}
    \includegraphics[width=\linewidth]{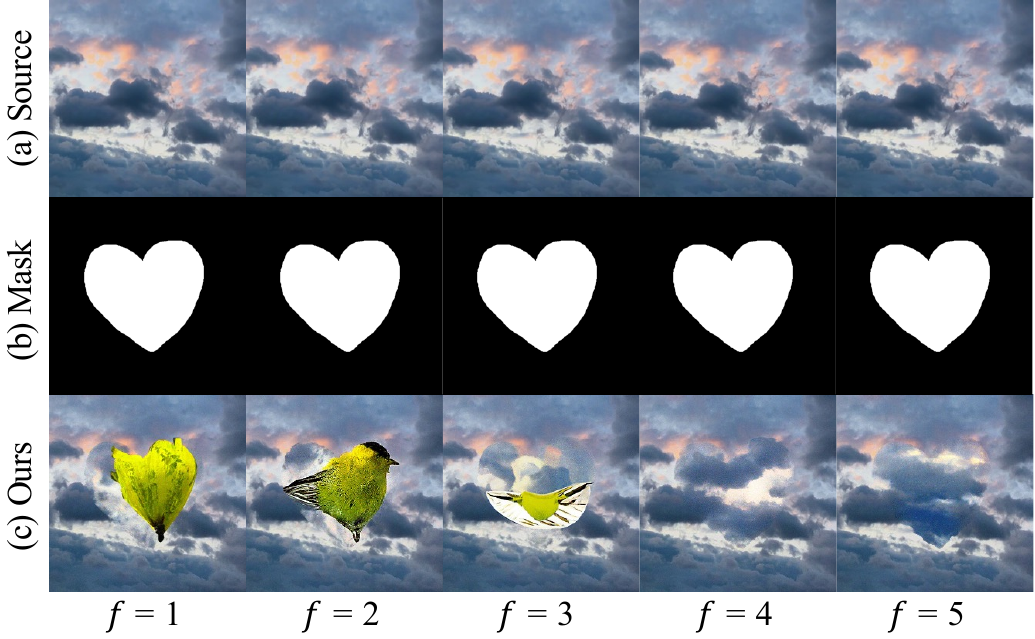}
    \end{center}
    \vspace{-1.2em}
    \caption{\textbf{Limitation of Soundini} Given bird song sound and heart-shape mask, Soundini fails to obtain natural videos associated with the beating of the wings of a bird.
 }
    \label{fig:limitation}
    \vspace{-1.5em}
\end{figure}

\vspace{-0.8em}
\section{Discussion}
\vspace{-0.8em}
\myparagraph{Limitation} Soundini successfully drives changes in visual style to be consistent with changes in sound. However, the performance of our framework degrades when the mask shape is not consistent with the content of the sound within the mask area. This is because our framework deals mainly with the pixel-wise relationship between the two adjacent frames~(see Fig.~\ref{fig:limitation}).

\myparagraph{Societal Impact} Soundini has the potential to raise ethical concerns about the generation of immoral content. As the user has complete control over the video editing and can apply styles that may contain inappropriate visual content, it is important to consider the social impact of such applications. In addition, inappropriate use of editing to create fake videos could have significant consequences for public figures, leading to the erosion of trust in the media.

\vspace{-0.8em}
\section{Conclusion}
\vspace{-0.8em}
Given sound and localization masks, we present a novel method to add realistic visual effects to videos using local sound guidance for diffusion. By leveraging audio-visual embedding space, our framework performs realistic video editing on various sound sources. Moreover, the optical flow guidance diffusion sampling mechanism is definitely helpful in achieving temporal consistency and producing context-adaptive results. We believe that our framework is the first work to address sound-guided video editing.

{\small
\bibliographystyle{ieee_fullname}
\bibliography{egbib}
}

\end{document}